\documentclass{article}
\usepackage{spconf,amsmath,graphicx}
\usepackage[]{algorithm2e}
\usepackage{booktabs}
\usepackage{listings}

\title{Improving End-to-end Speech Recognition with\\ Pronunciation-assisted Sub-word Modeling\thanks{The work reported here was conducted at the 2018 Frederick Jelinek Memorial Summer Workshop on Speech and Language Technologies, and supported by Johns Hopkins University with unrestricted gifts from Amazon, Facebook, Google, Microsoft and Mitsubishi Electric Research Laboratories.}}
\name{Hainan Xu, Shuoyang Ding, Shinji Watanabe}

\address{Center for Language and Speech Processing,\\
Johns Hopkins University,\\
3400 N. Charles St,\\
Baltimore MD, U.S.A.\\
{\texttt{\{hxu31,dings,shinjiw\}@jhu.edu}}}

\begin{document}
%\ninept
%
\maketitle
\begin{abstract}
Most end-to-end speech recognition systems model text directly as a sequence of characters or sub-words.
Current approaches to sub-word extraction only consider character sequence frequencies, which at times produce inferior sub-word segmentation that might lead to erroneous speech recognition output.
We propose \textit{pronunciation-assisted sub-word modeling} (PASM), a sub-word extraction method that leverages the pronunciation information of a word.
Experiments show that the proposed method can greatly
improve upon the character-based baseline, and also outperform commonly used
byte-pair encoding methods.
\end{abstract}
\section{Introduction}
\label{sec:intro}

In recent years, end-to-end models have become popular among the speech community.
Compared to hybrid-systems that consist of separate pronunciation, acoustic and language models, 
all of which need to be independently trained,
an end-to-end system is a single neural-network which implicitly models all three.
Although modular training of those components is possible \cite{chen2018modular}, an end-to-end model is usually jointly optimized during training. Among the different network typologies
for end-to-end systems, the attention-based encoder-decoder mechanism has proven to be very successful
in a number of tasks, including \textit{automatic speech recognition} (ASR) \cite{chiu2018state} \cite{kim2017joint} \cite{watanabe2017hybrid} and neural machine translation \cite{DBLP:conf/acl/SennrichHB16a}\cite{DBLP:conf/acl/Kudo18}.

Due to lack of a pronunciation dictionary, most end-to-end systems do not model
words directly, but instead model the output text sequence in finer units,
usually characters. This is one of the most attractive benefits of an end-to-end system as
it greatly reduce the complexities of overall architecture. 
However, it only works best for languages where
there is a strong link  between the spelling and the pronunciation, e.g. Spanish.
For languages like English, however, this approach might limit the performance of the system, especially
when there is no enough data for the system to learn all the subtleties in the language.
On the other hand, linguists have developed very sophisticated pronunciation dictionaries of high quality
for most languages, which can potentially improve the performance of end-to-end systems \cite{sainath2018no}. 

Sub-word representations have recently seen their success in ASR \cite{zeyer2018improved}.  % take care of this
Using sub-word features has a number of benefits for ASR, in that it can speed up both
training and inference, while helping the system better learn the pronunciation patterns of a language.
For example, if a sub-word algorithm segments the word ``thank'' into ``th-an-k'', 
this will make it easier for the ASR system to learn the association between the spelling ``th'' and the corresponding sound, which is not a concatenation of ``t'' and ``h''.
However, it should also be noted that lots of these methods are designed for
text processing tasks such as neural machine translation, and thus are only based on word spellings and
do not have access to pronunciation information. It is therefore possible for
these algorithms to break a word sequence into units that do not imply well-formed correspondence to phonetic units,
making it even more difficult to learn the mapping between phonemes and spellings.
For example, if a sub-word model sees a lot of ``hys'' in the data, it might process the word
``physics'' into ``p-hys-ics'', making the association with the ``f'' phoneme hard to learn.
We argue it is far from ideal to directly apply these methods to ASR and improvements
should be made to incorporate pronunciation information when determining sub-word segmentation.

This paper is an effort on this direction by utilizing a pronunciation
dictionary and an aligner. We call this method \textit{pronunciation-assisted sub-word modeling} (PASM), which adopts \texttt{fast\_align} \cite{dyer2013simple} to align a pronunciation lexicon file and use the result
to figure out common correspondence between sub-word units and phonetic units.
We then use the statistics collected from this correspondence to guide our segmentation process
such that it better caters to the need of ASR. The proposed method would work on a variety of
languages with known lexicon, and would also work in other tasks, e.g. speech translation.

This paper is organized as follows. In section \ref{prior}, we describe prior work;
in section \ref{method}, we give a detail description of our proposed method,
followed by section \ref{experiments}, where we report our experiment results.
We will conduct an analysis and discussion of the results in section \ref{discussion}
and then talk about future work in section \ref{future}.

\section{Related work}\label{prior}
The use of a pronunciation dictionary is the standard approach in hybrid speech recognition.
\cite{chen2015pronunciation} use the phone-level alignment to
generate a probabilistic lexicon and proposed a word-dependent silence model
to improve ASR accuracy; for use in end-to-end ASR models, \cite{sainath2018no} investigated
the value of a lexicon in end-to-end ASR.
Sub-word methods have a long history of application in a number of language related tasks.
%\cite{shinoda2000mdl} used \textit{minimal description length} as the criteria for sub-word modeling for speech recognition.
\cite{bulyko2012subword} used sub-words units in particular for detecting unseen words.
\cite{van2007vocabulary} used sub-words units in building
text-independent speech recognition systems.
\cite{smit2017improved} improved upon sub-word methods in WFST-based speech recognition.

Apart from the application in ASR, the most recent tide of adopting sub-word
representations is largely driven by neural machine translation.
\cite{DBLP:conf/acl/SennrichHB16a} proposed to use byte-pair encoding (BPE)
\cite{gage1994new} to build a sub-word dictionary by greedily keep the
most frequent co-occurring character sequences. Concurrently,
\cite{DBLP:journals/corr/WuSCLNMKCGMKSJL16} borrow the practice
in voice search \cite{DBLP:conf/icassp/SchusterN12} to segment words into
\textit{wordpiece} which maximizes the language model probability.
\cite{DBLP:conf/acl/Kudo18} augments the training data with sub-word segmentation
sampled from the segmentation lattice, thus increasing the robustness of
the system to segmentation ambiguities.

\section{Method} \label{method}
\subsection{Method Overview}
The high-level idea of our method is as follows: instead of generating a sub-word segmentation scheme by collecting spelling statistics from the tokenized text corpus,
we collect such statistics only from the \textit{consistent letter-phoneme pairs} extracted from a pronunciation lexicon.
The automatically extracted consistent letter-phoneme pairs can be treated as an induced explanation for the pronunciation of each word,
and hence, such pairs will ideally contain no letter sequences, i.e. sub-words, that will lead to ill configurations such as ``p-hys-ics''.

We generate sub-word segmentation schemes in 3 steps:
\begin{enumerate} \itemsep -2pt
    \item Using an aligner to generate a letter-phoneme alignment from a pronunciation dictionary
    \item Extract consistent letter-phoneme pairs from alignment 
    \item Collect letter-sequence statistics from the consistent letter-phoneme pairs
\end{enumerate}
To simplify the model and generalize to unseen words, we do not perform word-dependent sub-word modeling in this work. Our model generates a list of sub-words with weights, and we split
any word with those sub-words. 

\subsection{Method Description}
\subsubsection{Letter-phoneme Alignment Generation}
\begin{figure}
\begin{center}
\caption{A Simple Alignment for the Word ``SPEAK''\label{alignment}}
\includegraphics[scale=.4]{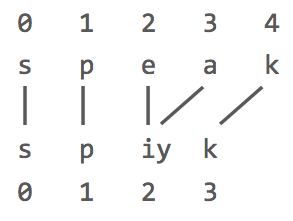}
\end{center}
\end{figure}

We use \texttt{fast\_align} to generate an alignment between letters and phonemes i.e. its pronunciation, which will be able to find common patterns of letter sequences that correspond to certain phonetic units.
%The output is represented as a set of (letter-index, phone-index) pairs.
For example, for the alignment shown in Figure \ref{alignment},
it is represented as a set,
$$
\{(0,0), (1,1), (2,2), (3,2), (4,3)\} 
$$
where each element in the set is a pair of (letter index, phone index), both being 0-based.
In this case, letters 2 and 3 are aligned to the same phoneme 2. In practice, we could have
one-to-one (e.g. ``cat''), one-to-many (e.g. ``ex''), many-to-one (e.g ``ah'') and even many-to-many alignments (linguistically this should not happen for most languages but this is a good indicator of an ``outlier'' case, e.g. a French word in an English corpus which the aligner does not know how to process properly). 

\subsubsection{Finding Consistent Letter-phoneme Pairs}
Formally, a consistent letter-phoneme pair $(L, P)$ is consisted of a letter sequence (or sub-word) $L = (l_1, ..., l_n)$ and a phoneme sequence $P = (p_1, ..., p_m)$. These pairs are heuristically extracted from the letter-phoneme alignment generated by \texttt{fast\_align}, and are then further refined to reduce noise mostly introduced by erroneous alignments.

\begin{figure}
\begin{center}
\caption{An Alignment with Crossovers\label{cross_over_figure}}
\includegraphics[scale=0.4]{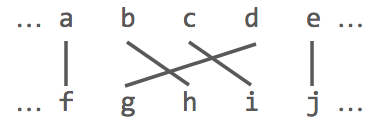}
\end{center}
\end{figure}

\textbf{Extraction} \quad As \texttt{fast\_align} is a re-parameterization of IBM model 2, a typical alignment method for statistical machine translation, it does not limit itself in generating
monotonic alignments. There could be cross-overs in its output, like in Figure \ref{cross_over_figure}, as well as ``null-alignments'', where a letter is aligned to a ``null'' symbol.

In the case of non-crossing alignments like the one shown in Figure \ref{alignment}, we simply extract each connected sub-sequences. The extracted consistent pairs of this example would be 
${(s, s)}, {(p, p)}, {(e\text{-}a, iy)},  {(k, k)}$. When there are cross-overs in the generated alignments, like in Figure \ref{cross_over_figure},
we take the maximum clustered sub-graph as a consistent pair, i.e. extracting 
${(b\text{-}c\text{-}d, g\text{-}h\text{-}i)}$. 

If a letter is aligned to a ``null'' symbol, we do not count this as a ``cross-over'' and keep the letter-to-null mapping for later processing.

\textbf{Refinement} \quad Refinement over the consistent letter-phoneme pairs is performed under the following criteria:
\begin{enumerate} \itemsep -2pt
    \item min-count constraint: $L$ must occur at least $N$ times in the training corpus,
    \item proportion constraint: of all the words containing $L$ in the corpus, at least a certain fraction $p$ of all occurrences is mapped to a particular phone-sequence $P$.
\end{enumerate}
% In practice we fix $N = 100$ and $p = 0.5$. % After processing we only keep the selected letter-sequences and their counts in data for future steps, and discard the pronunciations
% \footnote{In other words, in order for a sub-word to be selected, it must i) occur in the data at least $n > 100$ times, ii) has a corresponding phoneme sequence which occurs at least $n / 2$ times.}.
% The counts are based on the number of occurrences of the sub-words
% in the training corpus, and would act as weights of sub-word units, where
% higher counts indicate higher weights.

\subsubsection{Collecting Letter Sequence Statistics}
Recall that while we use pronunciation lexicon to extract consistent letter-phoneme pairs, our ultimate goal is to collect reliable statistics of the letter sequences (i.e. sub-word) to guide the sub-word segmentation process. Such statistics has nothing to do with phonemes, which means it needs to be marginalized. We perform the marginalization by summing up the counts of each type of letter sequence over all possible types of phoneme sequences. The marginalized counts would act as weights of sub-word units, where higher counts indicate higher weights.

\subsection{Text Processing}
As with all the sub-word modeling methods, our text processing step takes tokenized word sequences as input and segment them into sequences of sub-words. The segmentation process is essentially a search problem operating on the lattice of all possible sub-word segmentation schemes over the word-level input. This segmentation space is constrained by the complete set of sub-words in the segmentation scheme generated above, with hypothesis priorities assigned by the associated weight statistics, where sub-words with
higher weights would have higher priorities. For example, if both ``\texttt{ab}'' and ``\texttt{bc}'' are chosen as sub-words, and ``\texttt{ab}'' occurs more often than ``\texttt{bc}'' according to the statistics, then ``\texttt{abc}'' would be split as ``\texttt{ab c}'' instead of ``\texttt{a bc}''.

\section{Experiments} \label{experiments}
We conduct our experiment using the open-source end-to-end speech
recognition toolkit ESPnet \cite{watanabe2018espnet}. We report the ASR performance on the Wall Street Journal (WSJ) and LibriSpeech (100h) datasets. 
Our baseline is the standard character-based recipe, using bi-directional LSTMs with projection layers as the encoder, location-based attention, and LSTM decoder, with a CTC-weight of 0.5 during training \cite{watanabe2017hybrid}. To fully see the effect of sub-word methods, 
we do not perform language model rescoring but report the 1st pass numbers directly. 

\begin{table}[htp]
\centering
\caption{WER Results of BPE Systems on WSJ}
\begin{tabular}{lllll}
\toprule
\textbf{Num-BPEs} & \textbf{50}    &  \textbf{108} &  \textbf{200} &  \textbf{400}  \\ \midrule
dev93  &   20.7 &      \textbf{19.5} & 21.3 & 24.6 \\
eval92 &   15.2     &      \textbf{15.6} & 17.7 & 20.0 \\ \midrule
\end{tabular}
\label{bpe}
\end{table}

\begin{table}[htp]
\centering
\caption{WER Results on WSJ}
\vspace{0.1cm}
\begin{tabular}{llll}
\toprule
               & \textbf{Baseline} & \textbf{PASM} & \textbf{BPE}    \\ \midrule
dev93  & 20.7     &\textbf{18.5}  & 19.5  \\
eval92 & 15.2     &\textbf{14.3}  & 15.6 \\ \midrule
\end{tabular}
\label{wsj}
\end{table}

\begin{table}[htp]
\centering
\caption{WER Results on LibriSpeech}
\begin{tabular}{llll}
\toprule
           & \textbf{Baseline} & \textbf{BPE}  & \textbf{PASM} \\ 
    \midrule
dev-clean  & 23.8     & 29.5 & \textbf{21.4} \\
dev-other  & 52.8     & 53.1 & \textbf{50.7} \\
test-clean & 23.2     & 29.5 & \textbf{21.3} \\
test-other & 54.8     & 55.3 & \textbf{52.8} \\ 
\bottomrule
\end{tabular}
\label{lib}
\end{table}

\begin{table*}[htp]
\centering
\caption{Samples of Segmented Text Under the PASM Scheme and BPE Schemes with Various Vocabulary Sizes}
\begin{tabular}{l p{0.9\textwidth}}
\toprule
    \textbf{Scheme}       & \textbf{Text}  \\ \midrule
original  & \texttt{the sale of the hotels is part of holiday's strategy to sell off assets and concentrate on property management}      \\
PASM & \texttt{\_th e  \_s a le  \_o f  \_th e  \_h o t e l s  \_i s  \_p a r t  \_o f  \_h o l i d ay ' s  \_s t r a t e g y  \_t o  \_se ll  \_o ff  \_a ss e t s  \_a n d  \_c o n c e n t r a t e  \_o n  \_p r o p er ty  \_m a n a ge m e n t}     \\
BPE-108  & \texttt{\_the \_s al e \_of \_the \_ h o t e l s \_is \_p ar t \_of \_ h o l i d a y ' s \_st r ate g y \_to \_s e l l \_of f \_a s s e t s \_and \_con c en t r ate \_ on \_pro p er t y \_m an a g e m en t}     \\
BPE-200 & \texttt{\_the \_s al e \_of \_the \_ho t e l s \_is \_p ar t \_of \_ho l id ay ' s \_s t r ate g y \_to \_s e l l \_of f \_as s e t s \_and \_con c ent r ate \_on \_pro p er t y \_m an a ge me       nt} \\
BPE-400 & \texttt{\_the \_s al e \_of \_the \_ho t e l s \_is \_p ar t \_of \_ho l id ay ' s \_s t r ate g y \_to \_s e l l \_of f \_as s e t s \_and \_con c ent r ate \_on \_pro p er t y \_m an a ge ment}      \\

\bottomrule
\end{tabular}
\label{tab:sample}
\end{table*}

We also compare our systems with BPE baselines. The BPE procedure follows the algorithm described in \cite{DBLP:conf/acl/SennrichHB16a}. All the PASM segmentation schemes are trained using the lexicon included in its default recipe, and we use $N = 100$ and $p = 0.5$. All the other hyper-parameters are independently tuned.

For the WSJ setup, we have kept the number of sub-word units to be the
same in BPE and PASM systems (both = 108). The results are shown in Table \ref{wsj}, where
we report the word-error-rates on the dev93 and eval92 sets. We see that, the use of BPE improves dev93 performance but hurts performance on eval92. PASM method gives consistent
improvements in the 2 datasets.

We also report the more BPE results on WSJ, adjusting number of BPE units in Table \ref{bpe}. We can see that having more BPEs actually hurts the performance\footnote{The character-based baseline has a vocabulary-size of 50.}. This is likely because
of the limited data-size of WSJ, which makes it hard to learn reliable BPE units. 

In Table \ref{lib}, we report the WER results on the LibriSpeech dataset, using
the parameters described in \cite{zeyer2018improved}. We have seen that PASM significantly
improves the character-based baseline; BPEs do not help in this case, possibly due to
poor hyper-parameter tuning.

\section{Analysis} \label{discussion}

In Table \ref{tab:sample}, we show the output after the BPE procedure of the first sentence in the WSJ training
data, and compare that with the result of the PASM algorithm\footnote{For clear presentation,
we use the underline \_ character to represent a ``start-of-word'' symbol.}.

% Original text: \\
% \textit{the sale of the hotels is part of holiday's strategy to sell off assets and concentrate on property management}

% BPE (108): \\
% \textit{
% \_the \_s al e \_of \_the \_ h o t e l s \_is \_p ar t \_of \_ h o l i d a y ' s \_ st r ate g y \_to \_s e l l \_of f \_a s s e t s \_and \_con c en t r ate \_ on \_pro p er t y \_m an a g e m en t}

% BPE (400): \\
% \textit{
% \_the \_s al e \_of \_the \_ho t e l s \_is \_p ar t \_of \_ho l id ay ' s \_s t r ate g y \_to \_s e l l \_of f \_as s e t s \_and \_con c ent r ate \_on \_pro p er t y \_m an a ge ment}

% BPE (5000): \\
% \textit{
% \_the \_sale \_of \_the \_hotels \_is \_part \_of \_holiday ' s \_strategy \_to \_sell \_off \_assets \_and \_concentrat e \_on \_property \_management
% }

% PASM: \\
% \textit{
% \_th e  \_s a le  \_o f  \_th e  \_h o t e l s  \_i s  \_p a r t  \_o f  \_h o l i d ay ' s  \_s t r a t e g y  \_t o  \_se ll  \_o ff  \_a ss e t s  \_a n d  \_c o n c e n t r a t e  \_o n  \_p r o p er ty  \_m a n a ge m e n t}

From the examples above, we observe the following:
\begin{itemize}
    \item The PASM method correctly learns linguistic units, including ``le'',
    ``th'', ``ay'', ``ll'', ``ll'', ``ss'', ``ge'', which correspond to only one phoneme, but were not correctly handled in the BPE case.
	\item The BPE learns some non-linguistic but frequent-seen units in data, e.g. ``the'', ``ate''. In particular, the pronunciation associated with ``ate'' in the 2 occurrences are very different (concentr-ate vs str-ate-gy), which might make it harder for the system to learn the associations.
    \item As the number of BPE units increases, we see more sub-word units that do not conform to linguistic constraints, e.g. ``as-s-e-t-s'' and ``of-f'' in BPE-400. In this case, the 2nd ``s'' ``asset'' and 2nd ``f'' in ``off'' would have to be silent in terms of pronunciation,
which would likely confuse the training of end-to-end systems unless there is a huge amount of data. 
%This is in accordance with the results we have seen in Table \ref{bpe}, where more BPEs hurt performance.
\end{itemize}

%\subsection{Learned (end-to-end ASR) Alignments}

\section{Conclusion and Future Work}\label{future}
In this work, we propose a sub-word modeling method for end-to-end ASR based on
information from their pronunciations. Experiments show that the proposed method
gives substantial gains over the letter-based baseline, as measured by word-error-rates.
The method also outperforms BPE-based systems. We postulate that the
improvement comes from the fact that, the proposed method learns more phonetically meaningful
sub-words for speech tasks, unlike BPE which only take the spelling into consideration.

There are a lot of future work directions that we plan to explore. We will design
new algorithms for aligning pronunciation dictionaries that is tailored for speech tasks;
we will combine
the proposed method with BPE to further improve ASR performances and speed up systems;
we also plan to investigate
the application of the proposed method in hybrid ASR, machine translation, as well
as speech translation.

\label{sec:refs}

\bibliographystyle{IEEEbib}
\bibliography{refs}

\end{document}